\documentclass[a4paper]{article}

\usepackage{INTERSPEECH2021}
\usepackage{multirow}
\usepackage{tcolorbox}

\title{An Initial Investigation of Non-Native Spoken Question-Answering}
\name{Vatsal Raina, Mark J.F. Gales\thanks{
The research in this paper is supported by funding from the EPSRC DTP Studentship.
Thanks to Cambridge English Language Assessment for support and access to the Linguaskill data. The authors would like to thank members of the ALTA Speech Team for generating the ASR transcriptions for the responses.
}}
\address{
  ALTA Institute/Engineering Department, Cambridge University, UK}
\email{\{vr311,mjfg100\}@cam.ac.uk}

\begin{document}

\maketitle
\begin{abstract}

Text-based machine comprehension (MC) systems have a wide-range of applications, and standard corpora exist for developing and evaluating approaches. There has been far less research on spoken question answering (SQA) systems. The SQA task considered in this paper is to extract the answer from a candidate’s spoken response to a question in a prompt-response style language assessment test.  Applying these MC approaches to this SQA task rather than, for example, off-topic response detection provides far more detailed information that can be used for further downstream processing. One significant challenge is the lack of appropriately annotated speech corpora to train systems for this task. Hence, a transfer-learning style approach is adopted where a system trained on text-based MC is evaluated on an SQA task with non-native speakers. Mismatches must be considered between text documents and spoken responses; non-native spoken grammar and written grammar. In practical SQA, ASR systems are used, necessitating an investigation of the impact of ASR errors. We show that a simple text-based ELECTRA MC model trained on SQuAD2.0 transfers well for SQA. It is found that there is an approximately linear relationship between ASR errors and the SQA assessment scores but grammar mismatches have minimal impact.

\end{abstract}

\section{Introduction}
\label{sec:intro}

In recent years, extractive text-based machine comprehension (MC) has advanced significantly with many systems \cite{Zhang2020RetrospectiveRF, Lan2020ALBERT:, DBLP:conf/aaai/0001WZDZ020, yamada-etal-2020-luke, clark2020electra} outperforming human performance. The availability of standard corpora for reading comprehension has enabled these advances in text-based MC systems \cite{rajpurkar-etal-2016-squad, DBLP:conf/acl/RajpurkarJL18, herman-cnn, hill2015the, joshi-etal-2017-triviaqa}. In contrast, less work has been done for spoken question answering (SQA), where a machine gives an answer to a question from spoken content \cite{turmo-overview, Umbert2012FactoidQA, comas-sibyl}.

In this paper a form of machine listening comprehension is considered in the context of spoken language assessment where a candidate gives a spoken response to a prompt. The SQA model is required to extract the answer from the candidate's spoken response to the textual prompt. Automated assessment needs to ensure a candidate's response is relevant to the prompt \cite{raina-etal-2020-complementary, Wang2019}. MC would allow these systems to provide finer details than simple off-topic detection. Identifying the answer regions in a candidate's response can be used as additional features in downstream assessment tasks such as feedback.

The lack of appropriately annotated speech corpora in English poses a significant challenge to this SQA task. \cite{Tseng+2016} propose machine comprehension of spoken content using the TOEFL listening comprehension test as their corpus. Several deep models have been evaluated on this task: \cite{Tseng+2016} uses attention-based RNN, \cite{fang-hier} uses tree-structured RNN and \cite{chung-etal-2018-supervised} explores transfer learning from text-based question-answering. However, the TOEFL comprehension is a multiple-choice test that does not address the more challenging SQA task of extracting the answer region. \cite{Lee2018} propose Spoken SQuAD as an extraction-based SQA task. The Spoken SQuAD dataset takes the SQuAD MC dataset \cite{rajpurkar-etal-2016-squad} and passes the passages through a text-to-speech system to generate spoken equivalents. SpeechBERT \cite{Chuang2020} is an end-to-end SQA model trained on Spoken SQuAD. However, Spoken SQuAD is artificially constructed from text data, meaning it does not capture the elements of speech that distinguishes it from written text such as relaxed discourse, grammatical inconsistencies and disfluencies. \cite{odsqa} release a large scale SQA dataset with real speech but this dataset is in Chinese.

To address the lack of data, a transfer learning \cite{survey-pan} style approach is adopted. A system trained on a text-based MC task is directly evaluated on the SQA task with non-native spoken responses transcribed manually and with automatic speech recognition (ASR) systems. There are several mismatches that occur between training and evaluation: a domain shift from comprehension of text documents to spoken responses; grammatical irregularities not present in native speech due to non-native speakers of English; word errors due to ASR transcriptions.
In deployment of SQA models, ASR systems will inevitably be involved at the beginning of the pipeline. Therefore, it is important to explore the direct consequence of ASR errors on various automated assessment tasks.
The impact of these ASR errors on several spoken language assessment tasks has previously been evaluated \cite{Lu2019, Knill2018}. \cite{Lee2018} demonstrates that ASR errors have catastrophic impact on machine comprehension. This paper extends the investigation of the sensitivity of SQA systems to  ASR errors. The empirical relationship between ASR errors and SQA assessment scores is explored here. Simple data augmentation approaches are considered in order to bridge the differences between the text and speech domains.

\section{Text-Based Machine Comprehension}

Extractive reading comprehension tasks aim to locate the correct answer to a question in a context document. Hence, the task can be described by the set $\langle Q, P, A\rangle$, where $Q$ denotes the question, $P$ denotes the passage and $A$ denotes the answer span within the passage. Typical MC models take the question, $Q$, and the passage, $P$, as the input and return the start and end positions of the answer, $A$. Exact Match and F1 score are the typical performance metrics used for assessing text-based MC models.
The work in this paper focuses on MC models trained using SQuAD 2.0 \cite{DBLP:conf/acl/RajpurkarJL18}, the most recent version of the Stanford Question Answering Dataset. SQuAD 2.0 makes the task of extractive reading comprehension increasingly challenging as unanswerable questions are incorporated in the dataset. Hence, the MC models are required to abstain from giving an answer if it is not present in the passage.

State of the art for text-based MC is largely dominated by pre-trained language models (PrLMs) \cite{devlin-etal-2019-bert, liu2020roberta, NEURIPS2019_dc6a7e65, Lan2020ALBERT:, clark2020electra} on the encoder side based on the transformer architecture introduced by \cite{NIPS2017_vaswani}. ELECTRA-based models have been particularly successful on the SQuAD 2.0 leaderboard \cite{clark2020electra, Zhang2020RetrospectiveRF}.
Hence, the text-based MC model used in this paper is an ELECTRA encoder with a question-answering head as the decoder. The model returns separate probability distributions for the start and end positions over the sentence. An equivalent BERT model is considered too to contrast with the ELECTRA implementation.
During evaluation, the start and end positions are selected with the greatest probability sum. By default, the start and end positions of unanswerable questions are marked as 0.

\section{Spoken Question-Answering}

An SQA task is considered for which a system is required to extract the answer from a spoken passage to a textual or spoken question; a form of listening comprehension. In this paper, the focus is on textual questions.

The spoken content may be viewed from two perspectives: the original audio recording; or as transcribed speech using manual transcriptions or an ASR system. Hence, SQA is often evaluated using two metrics. \cite{Lee2018} propose the audio overlap score (AOS) which assesses the audio recording region. Alternatively the text overlap score (TOS) measures performance of the text overlap. This paper uses both metrics for assessing SQA models. In both cases, the overlap score is defined as the ratio between the intersection and the union of the prediction and the true answer regions.
$$\textrm{TOS} = \frac{\mathbf{x}\cap \mathbf{y}}{\mathbf{x}\cup \mathbf{y}}, \hspace{10pt} \textrm{AOS} = \frac{X\cap Y}{X\cup Y}.$$
$\mathbf{x}$, $\mathbf{y}$ denote the words of the predicted answer and the ground-truth answer respectively. $X$, $Y$ denote the audio segments of the predicted answer and the ground-truth answer respectively. The two performance metrics assess different aspects of an SQA system. TOS assesses the ability of the model to find \textit{what} is the answer while AOS informs the ability of the model to find \textit{where} is the answer.

This paper directly applies a system trained on text-based MC to SQA. The target domain is non-native speech as part of the responses from a prompt-response style language assessment test. The SQA system is required to find the answer to the prompt within the candidate's spoken response. Therefore, parallels can be drawn between the text-based MC task of $\langle \textrm{question}, \textrm{passage}, \textrm{answer} \rangle$ and the SQA task of $\langle \textrm{prompt}, \textrm{response}, \textrm{answer} \rangle$:
$$\textrm{question}\leftrightarrow{\textrm{prompt}}, \hspace{10pt} \textrm{passage}\leftrightarrow{\textrm{response}}.$$

The question-answering models are trained on the text-based MC dataset SQuAD2.0 and ported to the SQA task without any fine-tuning. Text predictions for the TOS are found using the transcriptions of the spoken response while the predicted audio segments for the AOS are found by forced alignment between the words in a transcription and the audio recording using a modified Damerau-Levenshtein algorithm \cite{knill19}.

The source and target domains differ in three respects. Speech and text data are inherently different meaning the text-based MC model is required to handle the natural disfluencies present in speech. As the SQA task is part of a candidate assessment of English learners, grammatical errors may be present as the candidates are non-native speakers; these errors will not be present in the text source domain. ASR is used to transcribe the responses in the target SQA domain, again these errors will not be present in the SQuAD2.0 data.

\section{Data Augmentation}

The MC models are trained in the source text domain but they are evaluated in the target speech domain with non-native speakers. The lack of speech data prohibits finetuning in the target domain. Data augmentation, where the training data undergoes some form of modification to create new examples, is readily used in low resource tasks (e.g. \cite{Cui2015DataAF, wei-zou-2019-eda}). Here, the purpose of data augmentation is to introduce disfluencies in the text domain such that the passages are closer in nature to the non-native spoken responses in the target domain. Two unstructured data augmentation approaches are considered: back-translation and text-to-speech transcriptions.

Back-translation is applied on the textual passages in SQuAD 2.0. A machine translation model translates a given passage to a foreign language and a reverse machine translation model translates back to English. In both directions, the maximum likelihood outputs are taken. The back-translated passage can expect to have similar meaning to the original passage but some disfluencies may be present
due to translation errors. The back-translated passage is used as an additional training example paired with the original question that allows the MC model to become more robust to disfluencies and hence improved transferability to non-native speech. Pre-trained machine translation models are publicly available.

Text-to-speech systems can be used to convert a textual passage from SQuAD 2.0 into a synthetic spoken equivalent. Passing the synthetic spoken passage through an ASR system returns a textual passage with ASR errors. The ASR errors can be seen to mimic the natural disfluencies present in the non-native speech domain. Therefore, text-to-speech transcriptions are treated as an additional data augmentation strategy to improve portability of MC models trained in the text domain and evaluated in the non-native speech domain.

\section{Experiments}

The text-based ELECTRA MC model was trained on SQuAD 2.0 \cite{DBLP:conf/acl/RajpurkarJL18}. This training dataset contains 87K answerable and 43K unanswerable questions. The performance of the ELECTRA model was evaluated on the SQuAD 2.0 validation set. The validation set consists of 5.9K answerable and 5.9K unanswerable questions. The ELECTRA MC model was trained for 2 epochs (4 hours) with hyper-parameters as specified in \cite{clark2020electra} \footnote{The ELECTRA-large from https://huggingface.co/transformers/ \\ model\_doc/electra.html} and results ensembled over 8 seeds on an nVidia GTX 980M graphics card. An equivalent trained BERT MC model \footnote{Trained model available at https://huggingface.co/deepset/\\ bert-large-uncased-whole-word-masking-squad2} was considered for comparison with ELECTRA. The implementation of the ELECTRA model in this paper \footnote{https://github.com/VatsalRaina/question\_answering\_squad2} achieves an EM of 86.9 and F1 score of 89.7 while the BERT model achieves an EM of 80.2 and F1 score of 83.2 on the validation set from SQuAD2.0.

The experiments for the target domain were run on prompt-response pairs from the Linguaskill-Business (L-Bus) Use of Business English test \footnote{https://www.cambridgeenglish.org/exams-and-tests/linguaskill/}. In this test, a candidate is required to provide spoken responses to prompts from five sections (A-E). For this paper, prompt-response pairs are only considered from section C, D and E that correspond to the more challenging long free-speaking parts of the test. Prompts for section C and D tend to be more open-ended e.g. ``\textit{talk about your colleague}" or ``\textit{describe graph X}" respectively, while section E prompts are more directed e.g. ``\textit{where can you get a loan}". The portability of text-based MC models was evaluated on test data from L-Bus, comprising 220 speakers across the CEFR grades A1-C2; mapping the grades to a six-point scale of 0-6, the average grade is 3.9 with a standard deviation of 1.2. Manual transcripts are available for the spoken responses, as well as ASR transcriptions. All utterances with ``unclear" words were removed. Audio time-stamp information was obtained by force-aligning the audio and manual transcriptions using a baseline ASR system. The manual transcriptions were annotated to indicate the regions of the response that answer the prompt. These answer regions were used as the ground-truth labels for the SQA task. Figure \ref{fig:example} depicts example annotations from questions in sections C and E of L-Bus.

\begin{figure}[t]
\begin{tcolorbox}
\begin{tabular}{@{}p{7cm}}
\textbf{Section C}: \\
\textit{Talk about why you like working with your colleague.} \\
``I like to work with him in project for new company he \underline{good relation with other people}". \\\\
\textbf{Section E}: \\
\textit{Where can you get a loan for your business?} \\
``Maybe he can go to make a good plan and then go to \underline{bank} he can try to get some finance over there".
\end{tabular}
\end{tcolorbox}
\caption{Example questions and non-native transcribed responses from sections C and E in L-Bus with annotated answer regions underlined.}
  \label{fig:example} 
\end{figure}

Table \ref{tab:stat} shows the statistics of the evaluation set for SQA. The level of grammatical errors, GEC error (\%), of the responses was calculated using the ERRANT system \cite{bryant-etal-2017-automatic}.

\begin{table}[th]
\caption{Average statistics with one standard deviation of L-Bus evaluation set for SQA.}
\centering
    \begin{tabular}{|l||c|c|c|}
    \hline
    & C & D & E    \\ \hline\hline
   \# samples & 141 & 146 & 660 \\ \hline
    \# prompt words & $30_{\pm 5.1}$ & $46_{\pm 6.5}$ & $11_{\pm 2.8}$  \\ 
    \# response words & $89_{\pm 32}$ & $93_{\pm 33}$ & $32_{\pm 12}$  \\ 
    \# answer words & $6.8_{\pm 9.8}$ & $23_{\pm 15}$ & $6.3_{\pm 5.9}$  \\ 
    response time (s) &  $53_{\pm 10}$ & $57_{\pm 7.0}$ & $17_{\pm 3.5}$  \\ 
    answer time (s) & $3.9_{\pm 4.3}$ & $15_{\pm 11}$ & $3.4_{\pm 3.1}$  \\ 
    GEC error (\%) & $13_{\pm 6.1}$ & $15_{\pm 6.7}$ & $13_{\pm 7.8}$  \\ \hline
    \end{tabular}
    \label{tab:stat}
\end{table}

The ensembled text-based ELECTRA MC model and the BERT MC model were applied directly to the text prompts and transcribed responses of sections C, D and E of the spoken language assessment task. The overlap scores are given in Table \ref{tab:cde} for the manual (MAN) transcription of the responses. The performance for section D is close to random while the text-based ELECTRA MC model performs best on section E of L-Bus which has prompts most similar to SQuAD2.0. It is clear that the open-ended nature of the questions in sections C and D, as seen in Figure \ref{fig:example}, make them less appropriate for this SQA task. The remainder of the results for the SQA focus on section E of L-Bus. As expected, the ELECTRA model outperforms BERT on L-BUS as it's better performing in the source domain task of SQuAD 2.0. Hence, the remainder of the discussion will focus on results with the ELECTRA model. The impact of the non-native speaker grammatical errors was then assessed. Note, the answer regions for grammatical error corrected (GEC) responses were re-annotated and only TOS can be computed. A small increase in TOS for GEC compared to MAN can be seen, despite a GEC word error rate of 13\% (Table \ref{tab:stat}, section E). This indicates that grammatical errors have small impact on the portability of the text-based MC model for this SQA task. This limited sensitivity may be due to using unigrams to calculate TOS coupled with the information-centric nature of question-answering.

\begin{table}[th]
\caption{TOS and AOS for manual (MAN) and grammatical errors corrected (GEC) transcriptions. Performance for text-to-speech (TTS) and machine-translation (MT) augmentation techniques.}
\centering
    \begin{tabular}{|lc||c||c|c|c|}
    \hline
    & & \footnotesize{SQuAD}& \multicolumn{3}{c|}{Linguaskill} \\
   & & \footnotesize{2.0} & C & D & E    \\ \hline\hline
      \multirow{2}*{BERT} & AOS & --- & 44.1 & 26.8 & 49.4 \\ 
                          & TOS & 90.3 & 48.0 & 28.6 & 49.4 \\\hline\hline
 \multirow{2}*{ELECTRA}  &  AOS & --- & 48.6 & 22.9 & 56.5 \\ 
        & TOS & 92.8 & 50.8 & 31.2 & 56.1 \\ \cline{2-6}
  +GEC  &  & ---  & --- & --- & 56.3 \\
  +MT  & TOS & 92.4 & 45.0 & 17.1 & 56.8 \\
  +TTS   &  & 92.4 & 47.0 & 18.6 & 56.6 \\
\hline
    \end{tabular}
    \label{tab:cde}
\end{table}

A second-set of human annotations were produced on a subset of L-BUS for SQA, 202 responses were randomly sampled from section E. The human annotations achieved a TOS of 64.9\% and an AOS of 68.3\% on the subset, indicating the challenges of annotating SQA for non-native speech. The ELECTRA model achieved a TOS of 58.2\% and an AOS of 58.6\% on this subset, which is considerably better than random.
Despite not achieving human performance, porting an MC system to this SQA task is a reasonable approach.

\begin{figure}[t]
\centering
    \includegraphics[width=1.0\linewidth]{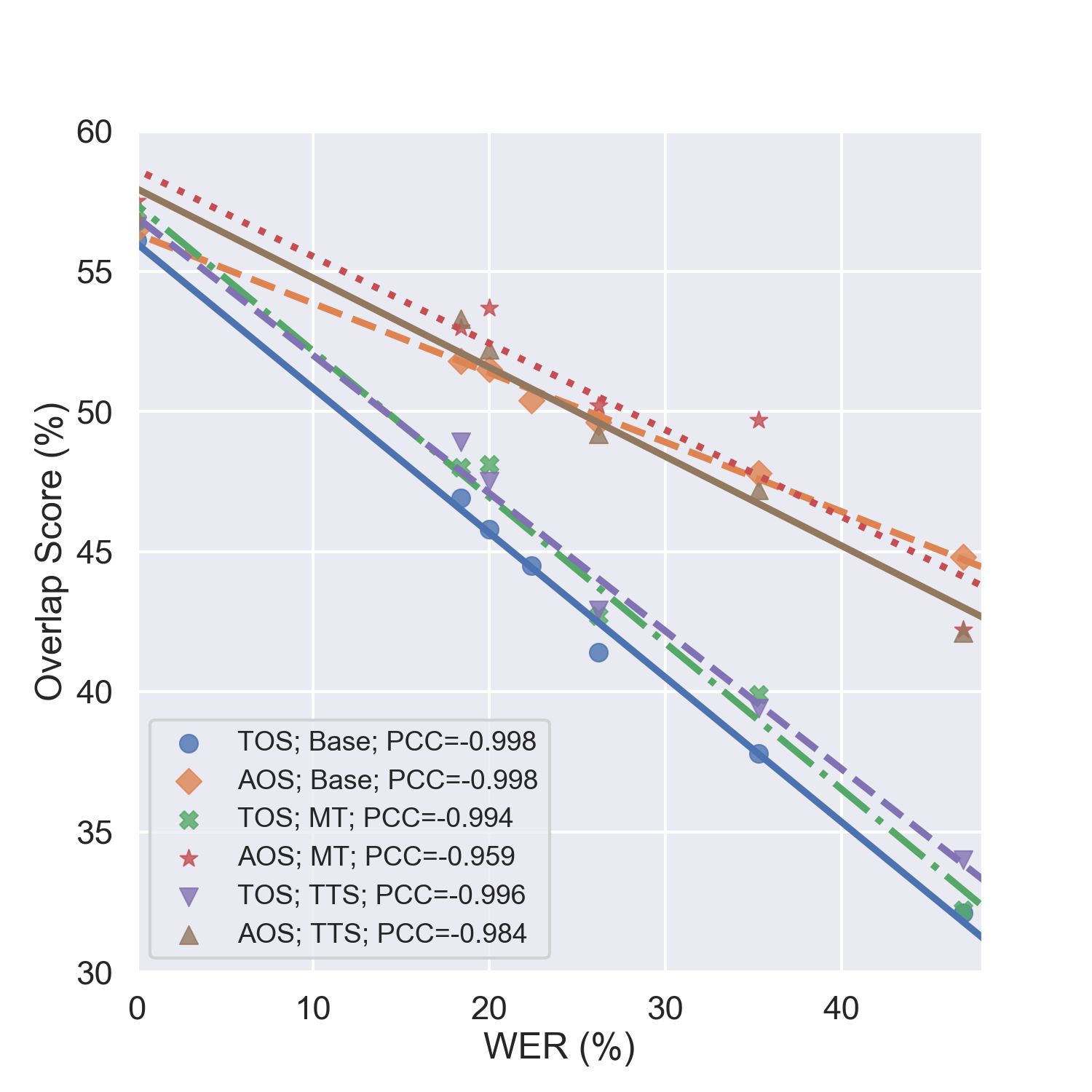}
  \caption{Impact of ASR errors on AOS and TOS on section E of L-Bus with baseline, text-to-speech (TTS) augmented and machine-translation (MT) augmented ELECTRA models.}
  \label{fig:asr} 
\end{figure}

In practical SQA, manual transcriptions will not be available for the spoken content. Therefore, it is useful to investigate the impact of the ASR errors on the system performance. Several ASR systems were considered all with the same decoding vocabulary, but differ in terms of the complexity of the acoustic and language models. Each is a hybrid deep learning-HMM graphemic system. The acoustic models are trained on non-native learner English speech from Linguaskill.
Table \ref{tab:asr} and Figure \ref{fig:asr} present the performance of the ELECTRA model against word error rate (WER) computed using the manual transcriptions.

Five ASR systems are considered (ASR1, ASR2, ASR3, ASR4, ASR5), where the WERs are calculated over the responses in section E of L-Bus. {\bf ASR5} is a Tandem acoustic model with a trigram language model (LM) \cite{Dalen2015AutomaticallyGL}. {\bf ASR4} uses DNN hybrid acoustic models with the same language model. {\bf ASR3} uses a stacked hybrid DNN and LSTM acoustic model (equivalent to ASR1 in \cite{Lu2019} and system 2 in \cite{Knill2018}). {\bf ASR2} is a sequence teacher-student trained lattice-free MMI (LF-MMI) factorised time-delay neural network system (TDNN) \cite{Povey2016PurelySN, Povey2018SemiOrthogonalLM, Wong2016SequenceST, Wang2018SequenceTT} with a succeeding word recurrent neural network LM (su-RNNLM) (the same as ASR3 from \cite{Lu2019}). {\bf ASR1} is a LF-MMI chain model with TDNN and CNN layers as well as SpecAug.

\begin{table}[th]
\caption{TOS and AOS for manual (MAN) and five ASR models on section E of L-BUS with baseline, text-to-speech (TTS) augmented and machine-translation (MT) augmented ELECTRA models.}
\centering
    \begin{tabular}{|l||c|ccccc|}
    \hline
      & \multicolumn{6}{c|}{Transcript} \\
      & \footnotesize{MAN} & \footnotesize{ASR1} & \footnotesize{ASR2} & \footnotesize{ASR3} & \footnotesize{ASR4} & \footnotesize{ASR5} \\ 
      \hline\hline
      {\bf WER} & --- & 18.4 & 20.0 & 26.2 & 35.3 & 46.9 \\
      \hline\hline
 {\bf TOS}  & 56.1 & 46.9 & 45.8 & 41.4 & 37.8 & 32.1 \\
 +MT & 56.8 & 48.0 & 48.1 & 42.7 & 39.9 & 32.2 \\
 +TTS & 56.6 & 48.9 & 47.5 & 42.9 & 39.4 & 34.0 \\
 \hline\hline
 {\bf AOS} & 56.5 & 51.8 & 51.5 & 49.6 & 47.8 & 44.8 \\
+MT  & 57.5 & 53.0 & 53.7 & 50.2 & 49.7 & 42.2 \\
+TTS & 56.9 & 53.3 & 52.2 & 49.2 & 47.2 & 42.1 \\
\hline
    \end{tabular}

    \label{tab:asr}
\end{table}

TOS is significantly more sensitive to ASR errors than AOS. Both these scores are calculated from the ratio between two quantities: the amount shared between the true answer and the prediction (intersection), and the total unique span of the true answer and prediction combined (union). In terms of text, the intersection represents the number of shared words and the union represents the total number of unique words. For audio, the intersection and union relate to the  shared and total time segments respectively. AOS is not impacted by content, only by the position of the predicted span. The ASR errors make question-answering more challenging which leads to increased misalignment between predicted regions and true answer regions. Hence, it is expected that AOS will decrease with ASR WER. It is interesting that this performance degradation is linear for AOS, about 0.25\% for every percentage point increase in WER. Conversely, TOS is impacted by both content and the position of the predicted span. Every ASR error within the answer region impacts the TOS words' intersection and increments the words' union. Hence, the TOS will degrade faster than AOS. This is observed in Figure \ref{fig:asr} where for each percentage point increase in WER TOS  decreases by 0.52\%. It is interesting that, despite the complex interactions between ASR errors and the overlap scores, both TOS and AOS degrade linearly with increasing WER.


Unstructured data augmentation approaches were applied to the ELECTRA model to see if they improved transferability of the MC model to the SQA task. Back-translation augmentation of the SQuAD 2.0 data was performed using two languages and hence 4 trained machine translation models were used \footnote{ https://huggingface.co/Helsinki-NLP/opus-mt-en-fr \\ https://huggingface.co/Helsinki-NLP/opus-mt-fr-en https://huggingface.co/Helsinki-NLP/opus-mt-en-de \\ https://huggingface.co/Helsinki-NLP/opus-mt-de-en}. Text-to-speech transcriptions of the SQuAD data are made available by \cite{Lee2018} using an ASR system with a WER of 44\%. For the augmentation, the ELECTRA model is trained on both the original and augmented data.

Figure \ref{fig:asr} shows the effect of the augmented ELECTRA models on the assessment scores of the SQA task for manual transcriptions and ASR transcriptions with WERs up to 47\%. Table \ref{tab:asr} contrasts the TOS and AOS scores with the baseline models.
The TOS is linearly improved with the back-translation and the transcribed text-to-speech augmentation approaches. AOS is significantly improved with the back-translation augmentation and text-to-speech augmentation for small WERs but struggles to improve over baseline results for high WERs. A possible reason for reduced boosting of AOS for high WERs is that both augmentation strategies have been performed on text data and hence are not robust to large word error rates and so the model struggles to find the answer region in noisy responses. Both augmentation approaches are mildly effective at improving transferability between the source and target domains as the corruption of the textual training data helps to mimic the natural disfluencies present in non-native speech. It is evident that the improvements brought about by the augmentation strategies are in the speech domain and not in the text domain as from Table \ref{tab:cde}, the augmentation strategies do not improve the baseline TOS scores on SQuAD 2.0.

\section{Conclusions}

This paper has examined the impact of ASR errors on ported text-based machine comprehension model for a non-native spoken question answering task. The transferred model was evaluated on a free-speaking section of an English assessment test, Linguaskill Business. It was required to extract the answer to a question from a candidate's spoken response. Two standard metrics were used to assess the system performance: text overlap score and audio overlap score. It was found that an ELECTRA text-based model trained on the SQuAD 2.0 machine comprehension task transfers well to this SQA task. Grammatical errors have a low impact on the transferability from text to speech domains for question-answering related tasks. However, speech recognition errors introduced do impact performance. Interestingly, both text and audio overlap performance metrics degraded in a linear fashion with increasing WER. Unstructured augmentation approaches introduced disfluencies in the text data that were found to be effective in boosting performance on the SQA task across a diverse range of ASR WERs.

\bibliographystyle{IEEEtran}

\bibliography{mybib}

\end{document}